\documentclass{article}
\pdfoutput=1
\usepackage{geometry}
\usepackage{amsfonts}
\usepackage{amssymb}
\usepackage{amsthm}
\usepackage{graphicx}
\usepackage{subfig} 
\usepackage{natbib}
\setcitestyle{authoryear,open={(},close={)}}
\begin{document}
\title{Machine Learning and Knowledge: Why Robustness Matters}
\author{Jonathan Vandenburgh}
\date{}
\maketitle

\begin{abstract}
Trusting machine learning algorithms requires having confidence in their outputs. Confidence is typically interpreted in terms of model reliability, where a model is reliable if it produces a high proportion of correct outputs. However, model reliability does not address concerns about the robustness of machine learning models, such as models relying on the wrong features or variations in performance based on context. I argue that the epistemic dimension of trust can instead be understood through the concept of knowledge, where the trustworthiness of an algorithm depends on whether its users are in the position to know that its outputs are correct. Knowledge requires beliefs to be formed for the right reasons and to be robust to error, so machine learning algorithms can only provide knowledge if they work well across counterfactual scenarios and if they make decisions based on the right features. This, I argue, can explain why we should care about model properties like interpretability, causal shortcut independence, and distribution shift robustness even if such properties are not required for model reliability.
\end{abstract}

Suppose, following \cite{Degrave}, that a hospital trains a machine learning model to detect whether a patient has Covid-19 based on chest x-ray images. The hospital trains this model with publicly available data, including Covid-negative x-rays from an NIH dataset and Covid-positive x-rays from a dataset found on Github, and achieves excellent performance when testing the model. However, unbeknownst to the hospital, part of the reason why the model appears successful is that it has learned to discern differences other than Covid status between the images in the NIH dataset and the Github dataset. For example, x-rays in the NIH dataset of Covid-positive patients tend to have more space above a patient's shoulders, and by learning this correlation, the machine learning model is able to improve performance in identifying Covid in the training environment.

Even though the evidence available to the hospital suggests that the model works incredibly well, and even though it may work reasonably well in practice, it seems inappropriate for the hospital to rely on the algorithm for Covid diagnosis. Some guidance as to why this model is untrustworthy can be found in work on the robustness of machine learning models. First, it does not generalize well: while it performs well in the training context, it performs less well in a new environment where the data looks different than it does in training. Second, it fails an interpretability test: when using an algorithm to interpret how the model predicts Covid status, the model appears to be focused on irrelevant parts of the x-ray image, such as the space above a patient's shoulders. Third, it relies on a shortcut or confounder, making a decision based on a feature which is spuriously correlated with Covid status in the training data rather than based on features that are causally relevant to having Covid.

While these seem like compelling reasons not to trust the Covid detection algorithm described above, it is unclear how well these reasons generalize or how they fit into a broader conception of trustworthy machine learning. For example, some authors argue that interpretability is not necessary for trusting algorithms, since it is either very difficult or impossible to meaningfully interpret the inner workings of a ``black box" machine learning algorithm \citep{Duran, Krishnan}. Further, looking toward the dominant epistemic theory associated with trustworthiness, computational reliabilism \citep{Duran, Duran1}, the above factors are either irrelevant for trustworthiness or relevant only insofar as they are indicators of reliable performance \citep{Mishra}.

Another avenue for addressing the trustworthiness of machine learning models is to focus on the ethical implications these models have for those affected. When an algorithm makes a decision that affects important aspects of peoples' lives, such as medical care, it is reasonable to expect that the algorithm make a decision based on the right features relevant to each individual case and that there is transparency and accountability in how the decision is made \citep{Smart, Vredenburgh}. When such conditions fail, algorithms can make decisions that not only risk being incorrect, but that violate our moral expectations about how life-altering decisions should be made. However, invoking ethical considerations for trust cannot obviate the need for an epistemic account of trustworthiness, since a central source of moral criticism of algorithmic decisions stems from epistemic criticism of how well the algorithm works.

In this paper, I argue for a stronger epistemic condition on trustworthiness than computational reliabilism that captures the intuition that machine learning algorithms should be robust to changes to the data distribution, interpretable, and grounded in the right causal features. In particular, I will argue that a machine learning algorithm is trustworthy only if its users are in the position to know that its outputs are correct. Since knowledge requires that one's belief should be robust to counterfactual changes and should be grounded in the right reasons, this account can naturally explain why we want machine learning algorithms to get the right answer robustly and based on the right features. Furthermore, since the robustness requirement of knowledge can be made more precise through conditions like safety \citep{Sosa, Williamson, Pritchard}, this account can provide some guidance for evaluating algorithmic robustness.

The paper proceeds as follows. In \S1, I introduce algorithmic evidence, the question of trustworthiness, and reliabilism about machine learning. In \S2, I argue that machine learning models, while reliable, can fail to provide knowledge, offering comparisons to Gettier cases and belief on the basis of statistical evidence. In \S3, I introduce the safety condition on knowledge and how it can be interpreted in the context of machine learning. In \S4, I argue that model interpretation and causal shortcut detection are closely related to the safety condition, especially when used for counterfactual robustness analysis. In \S5, I argue that concerns about robustness to distribution shift and adversarial robustness can also be understood through the lens of safety, comparing these cases to similar examples in the philosophy literature.

\section{Algorithmic Evidence and Reliability}

Machine learning models have shown remarkable promise in classification tasks, where they can be used to classify animals, plants, and other common objects, detect disease, and identify sentiment from text. For example, machine learning models can detect breast cancer from images of cells with up to 95\% accuracy \citep{Benhammou} and can classify images containing 1000 different classes of objects, from academic robes to wombats, with outputs that are around 90\% correct \citep{Vasudevan}. In addition to classification, machine learning models can develop strategies for playing games like Go \citep{Silver} and can predict optimal taxation strategies in multi-agent simulations \citep{Zheng}. More recently, generative machine learning models can answer questions in natural language, as in the GPT models \citep{Brown}, and can generate images from text inputs \citep{Rombach}. 

From the perspective of the user of a machine learning algorithm, the output of an algorithm offers evidence informing one's beliefs. For example, an algorithm detecting breast cancer in an image taken from a biopsy offers evidence in favor of the patient having cancer. Users may utilize this evidence in different ways, ranging from using the algorithm's output as one factor in decision-making to fully relying on the algorithm to make decisions autonomously.\footnote{While I will not discuss practical considerations about the best way to utilize algorithmic evidence, \cite{Grote, Grote1} offer some epistemic and ethical considerations on this issue.} I take it that trusting an algorithm amounts to treating evidence from the algorithm as sufficient for belief: trusting a cancer detector involves accepting the algorithm's verdicts as correct, and trusting a large language model involves accepting the statements of the model as true.\footnote{Note that this only captures the epistemic dimension of trust and that trust is often thought to have a moral dimension, such as requiring confidence in the goodwill of the trusted party \citep{Baier}. While the moral dimension of trust raises interesting questions about whether computational systems can even be proper objects of trust \citep{Coeckelbergh, Ryan}, it can also highlight some interesting further properties of trustworthiness, such as restrictions on reward functions or designer intentions to rule out, say, an algorithm designed to maximize hospital profit that happens to provide epistemically trustworthy diagnoses.}

Algorithmic evidence offers a potentially new kind of evidence supporting belief. Some closely related kinds of evidence are outputs from scientific instruments, such as temperature readings from thermometers, and human testimony \citep{Goldberg1, Wheeler, Duede}. Similar to scientific instruments, machine learning algorithms use computational methods for translating multidimensional inputs (such as a gas sample or a list of patient features) into a single output humans can understand (a temperature reading or a diagnosis). Similar to human testimony, machine learning algorithms can communicate information (e.g., that a patient has a certain disease) based on complex and opaque reasoning from its environment. However, neither analogy is perfect: machine learning algorithms do not use transparent and well-understood computations like most scientific instruments and cannot be held accountable for their own conclusions as in the standard social practice of testimony.

A natural approach to epistemically evaluating machine learning algorithms is to look at how reliable the algorithms are in producing true beliefs. This appeals to Alvin Goldman's (\citeyear{Goldman2}) reliabilist theory of epistemic justification, where a belief is justified if it is produced through a reliable process, or a process which has a high probability of producing true beliefs. This theory is appealing in the context of machine learning because reliability closely tracks model accuracy: when forming beliefs on the basis of a machine learning algorithm, the probability of forming a true belief is equal to the accuracy of the model over the relevant set of situations. Furthermore, reliabilism is often invoked when epistemically evaluating testimony and scientific instruments and is a plausible account of the epistemic dimension of trust \citep{Goldberg2}. 

Following reliabilism, it seems that in many cases of highly accurate machine learning algorithms, we are justified in believing that their outputs are correct. If a machine learning algorithm is over 90\% accurate, as many of them are, then believing on the basis of the model has over a 90\% chance of producing true beliefs. While this seems to render many beliefs on the basis of machine learning algorithms justified, advocates of computational reliabilism recommend caution in drawing this conclusion. This is because justification depends on reliability in the context in which an algorithm is deployed, which often falls below the level of accuracy exhibited on curated testing datasets. For this reason, good training and test performance may be insufficient to conclude that an algorithm is reliable, which may require evidence like a history of successful implementation, evidence of robustness, or support from domain experts \citep{Duran}. 

Despite such caution, the high accuracy achieved by machine learning algorithms suggests that reliable algorithms are either already here or just around the corner. However, many people, including machine learning researchers and practitioners, are skeptical of trusting machine learning algorithms. Machine learning models are notorious for making mistakes that are obviously avoidable for humans, such as mistaking dough for a jigsaw puzzle \citep{Vasudevan}. Furthermore, machine learning models are ``black boxes" that make decisions through processes we can never fully understand, and our best attempts to interpret them often show models to be sensitive to features that ought to be irrelevant to the problem at hand, like the amount of space above a patient's shoulder when detecting Covid. While reliabilists may be inclined to dismiss such concerns provided that they do not challenge real-world accuracy, I will argue that these concerns pose genuine epistemic challenges to trusting machine learning algorithms by invoking the philosophical concept of knowledge.

\section{From Justification to Knowledge}

It is plausible that believing on the basis of machine learning models reliably leads to true beliefs. This would mean that one is justified in believing that the outputs of a model are true and that, in most cases, one's beliefs will be correct.\footnote{While I have focused on reliabilist accounts of justification in line with most work on the epistemology of machine learning, it is also plausible that believing machine learning models is justified on an internalist or evidentialist approach to justification \citep{Conee}. The output of an accurate machine learning model provides evidence supporting the truth of the output, and when one has no evidence conflicting with the output, believing in accordance with the model seems to fit one's evidence.} However, one of the central insights from recent epistemology is that having a justified true belief is insufficient to know that something is true \citep{Gettier, Zagzebski}. This is because a belief can be justified and true, but be true for a reason disconnected from one's justification. This renders one's account of why the belief is true unsatisfactory and makes the truth of one's belief the result of chance rather than good reasoning.

Cases illustrating this are typically called Gettier cases, following \cite{Gettier}. A standard example comes from \cite{Russell} and follows the expression ``a stopped clock is right twice a day." When seeing a stopped clock that displays 8:14, one is justified in believing that the time is 8:14 because the clock reading is good evidence and a reliable indicator that the time is as displayed. Furthermore, if by chance the time happens to be 8:14 as displayed on the stopped clock, this belief is correct. However, the process used to determine the time is completely disconnected from the actual time and only succeeds because, by sheer luck, the actual time lines up with the displayed time. Such a method produces the right answer for the wrong reason, with success depending on luck in the environment, and is an unsatisfactory way of forming beliefs. A standard account of what is missing in this case is knowledge: one may have a justified true belief based on a stopped clock, but one cannot know the time on that basis. This case is illustrative of the concept of knowledge in general, which requires having the right answer for the right reason, or for a reason that is connected to the truth in the right way, rather than because of lucky circumstances.

One might think that true belief on the basis of reliable machine learning algorithms resembles a Gettier case. For example, suppose that one has a reliable version of the Covid detection algorithm discussed in the introduction, which, despite its high accuracy, relies in part on the space above a patient's shoulders. This algorithm's output, and therefore belief on the basis of this output, is based on a feature unrelated to the presence of Covid and is therefore dependent on luck in how the patient is positioned. Thus, while this algorithm may be a reliable source of true beliefs, beliefs formed on the basis of it may not be connected to the truth in the right way, so the algorithm may not be able to confer knowledge.

One difference between this case and typical Gettier cases is that the dependence on luck is part of the belief-forming process rather than the external environment. In the stopped clock case, the process of forming beliefs based on clock displays is clearly reliable, but one fails to achieve knowledge because of uncontrollable bad luck in one's environment (i.e., the broken clock). On the other hand, when one relies on a machine learning algorithm that depends on the wrong features, the dependence on luck is built into the belief-forming process itself. This may make such cases more similar to another example of justified true belief that falls short of knowledge: belief based on purely statistical evidence. A popular example of belief based on purely statistical evidence is the Blue Bus case, which involves a hypothetical court case where one is trying to determine whether a bus company (referred to as the Blue Bus company) can be held liable for causing a bus crash simply because it operates 80\% of buses in a city \citep{Thomson, Wells, Enoch}. Reliabilism seems to support forming beliefs on the basis of such statistical evidence, because such a process will produce on average many more true beliefs than false beliefs. However, one cannot have knowledge on the basis of purely statistical evidence, because the evidence is not directly connected to the relevant proposition: statistics about bus operations do not directly bear on who is responsible for a crash. 

There are two further reasons for thinking that belief on the basis of statistical evidence is a good analogy for at least some cases of belief on the basis of a reliable machine learning algorithm. First, the reliability of some machine learning algorithms may be virtually guaranteed by the background statistics of the domain of application. For example, if one is detecting a disease that is found in 2\% of patients, then any machine learning model that rarely detects the disease will reliably produce true beliefs, even if the outputs have little or no connection to the presence of the disease. Second, machine learning models are generally trained to maximize accuracy rather than to develop a more sophisticated (e.g., causal) understanding of the task, risking the development of models that systematically have high accuracy based on the wrong features and therefore can only provide statistical evidence rather than knowledge \citep{Pearl}.\footnote{Note, however, that dependence on the right features and other properties associated with causal understanding may emerge from maximizing accuracy, at least in some contexts. See, e.g., \cite{Geiger}.}

The analogy with belief based on statistical evidence is also helpful for illustrating why knowledge may be necessary for trustworthiness beyond mere reliability. In the Blue Bus case, the statistical evidence that the Blue Bus company was most likely responsible is insufficient to stop inquiring into the case (for example, by seeking out eyewitness testimony) and is insufficient to punish the company for the crash. A common explanation for this is that knowledge is required before we can stop inquiring about whether something is true and before we can act on the basis of something \citep{Hawthorne, Kelp}. Analogously, for machine learning algorithms to deserve the kind of trust where we can believe the outputs without further inquiry and where we can use the outputs as the basis for action, we must know that the outputs are correct. This is related to the claim that knowledge is the norm of belief, meaning that it is appropriate to believe something only when we know that it is true \citep{Williamson, Smithies, Littlejohn}. If we accept the knowledge norm for belief as well as the claim that trustworthiness requires being in the position to believe appropriately, then a machine learning algorithm (or any other source of belief) is only trustworthy if we are in the position to know that its outputs are true. Thus, if it turns out that believing the outputs of a machine learning algorithm resembles a Gettier case or belief based on mere statistical evidence, providing justification but not knowledge, then we ought not trust these algorithms.\footnote{Note that this argument also extends to the epistemic dimension of trust in other people, which is often discussed in terms of reliability \citep{Goldberg2}. For example, I take it that, for epistemic reasons, one ought not trust people who say things they know to be true 90\% of the time but intentionally make up facts 10\% of the time or people who routinely say things based on statistical evidence (i.e., someone who assumes that jobs are held by people in the social demographic most likely to occupy that job).}

\section{Evaluating Knowledge: The Safety Condition}

The above discussion motivates the possibility that machine learning algorithms grant justification but not knowledge, providing the right answer for the wrong reason. One indication of this is that interpreting algorithms can highlight unexpected features apparently underlying the algorithm's decision, such as facts about a patient's positioning when detecting disease, which indicates reliance on the wrong features without necessarily leading to unreliable performance. However, interpretability may be a misleading consideration, as it is typically unclear how highlighted features factor into a model's decision, and one may think that the fundamental opacity of black box algorithms should lead to suspicion about drawing hasty conclusions from attempts at interpretation \citep{Duran, Krishnan}. One of the virtues of reliabilism is that it can reduce the challenge of inspecting a model's reasoning process to the measurable question of reliability: bad interpretability results are problematic only insofar as they lead to unreliability in practice \citep{Mishra}.

However, we can develop more sophisticated arguments about the possibility of knowledge from machine learning algorithms by turning to some of the tools developed in the theory of knowledge. A popular account of the difference between justification and knowledge invokes robustness to error. The motivating idea is that, while justification requires error to be unlikely overall, knowledge requires that each individual judgment also be robust to counterfactual errors. Counterfactual scenarios are modeled as nearby possible alternatives (or nearby possible worlds), which are different ways the world could have been that are relevantly similar to how things actually are. The framework of possible worlds was introduced to understand possibility and necessity \citep{Kripke} and has been applied to counterfactual reasoning \citep{Stalnaker, Lewis1} and the nature of meaning and identity \citep{Putnam, Kripke1}, serving as a fundamental tool for reasoning about alternative possibilities.

A popular condition on knowledge in this vein is the safety condition, which requires that one's beliefs could not easily have been wrong \citep{Sosa, Williamson, Pritchard}.\footnote{Similar theories designed to capture robustness to error include the sensitivity condition \citep{Nozick, Enoch} and David Lewis's (\citeyear{Lewis}) relevant alternatives theory of knowledge.} In terms of possible worlds, a belief is safe if, in nearby possible worlds where one holds the same belief, that belief is true. For example, when one believes the correct time based on a stopped clock, one's belief could have easily been wrong if the time had been different. In other words, the situation where one's belief is correct at 8:14 is similar to the situation where it is 8:15 but the clock displays 8:14, meaning that there is a similar scenario where one would have falsely believed that the time is 8:14, showing that the belief is not safe. Similarly, when one correctly believes that the Blue Bus company caused a crash because it operates most buses in the city, there is a similar hypothetical situation where a different company caused the crash, but where one still believes the Blue Bus company is responsible based on the statistical evidence. The safety condition works in these cases because the belief-forming mechanism is not appropriately connected to reality, so when considering counterfactual changes to the world, beliefs do not change as expected, leading to nearby worlds where the belief is false.

For belief on the basis of a machine learning algorithm, safety requires that the belief be true in similar worlds where the same belief is formed, which are just the similar worlds where the algorithm yields the same result. Whether machine learning algorithms can support safe belief depends on how similarity between possible worlds is analyzed. Typically, possible worlds represent more or less complete ways of describing a situation and similarity refers to a distance measure on the set of possible worlds. Thus, for any worlds $w_1$ and $w_2$ we can calculate the distance $d(w_1,w_2)$ between them, and the similar worlds to $w$ are those that are sufficiently close to $w$ according to $d$, say, all $w'$ such that $d(w,w')<\epsilon$. Given that machine learning involves representing the world in mathematical ways amenable to defining similarity measures, counterfactual theories seem particularly promising for investigating whether machine learning outputs are robust enough to support knowledge.

A natural first thought is that, in domains where one can easily define a distance measure over data points, one can use this distance measure to determine the distance between possible worlds. For example, in image classification, images can be represented as matrices of RGB color values and one can define the distance between two images as, say, the average of the distances between pixel values. However, this notion of similarity in the mathematical representation does not always correspond to intuitions of similarity between possible worlds. For example, in the case of Covid detection based on x-rays, the possible world where a patient was positioned a few centimeters differently is intuitively very close to the actual world, but the mathematical distance between images may be very high since the pixels of the same color would not be aligned across images.

Furthermore, utilizing a distance measure defined on a mathematical representation of a situation does not account for possible changes to variables left out of the mathematical representation. Consider two possible worlds where an algorithm detects Covid based on an image $I$: in the first world, $I$ is the image produced by a chest x-ray of a patient with Covid, but in the second world, the patient does not have Covid and instead someone hacked  the hospital's record system and made the patient's x-ray look exactly like $I$. Since the data representations are identical, any distance measure over the representation space judges the distance between the two situations as zero. However, conceived of as possible worlds, the two situations are very far apart: many facts would have to change about the first world to bring about the second world. 

This discussion suggests that the philosophical notion of distance between possible worlds underlying the safety condition for knowledge cannot be immediately reduced to a technical conception of distance in machine learning. While formalizing this notion in the context of machine learning would be a worthwhile project, evaluating such a formalization requires comparison to prior philosophical judgments of safety. The goal of the rest of this paper is to argue for some judgments about what is and is not required for machine learning algorithms to support safe belief. I will consider four robustness properties often demanded of machine learning algorithms: interpretability, independence from causal shortcuts, robustness to distribution shifts, and adversarial robustness.

\section{Interpretability and Causal Shortcuts}

One of the key properties differentiating knowledge from justified true belief is that knowledge involves getting the right answer for the right reason. This is also one of the goals of interpreting machine learning algorithms, where one searches for evidence that a model made its decision based on the right input features. A common method for interpreting algorithms involves generating local explanations, where for any individual data point, one can estimate the features of that data point that were most significant for the model's prediction.\footnote{Examples of algorithms for local explanations include LIME \citep{Ribeiro}, SHAP \citep{Lundberg}, and Grad-CAM \citep{Selvaraju}.} For example, in object classification in images, local explanation algorithms can highlight the part of the image estimated to have the largest impact on the prediction. Thus, when looking at a Covid detector's diagnosis for a particular chest x-ray, one wants the local explanation to highlight the patient's lungs rather than regions outside of the patient's body as an indicator that the model is using the right features.

However, given the black box nature of machine learning algorithms, interpretations may not adequately capture how a model makes decisions, and users may have a hard time judging whether an interpretation highlights the ``right features."\footnote{For an approach to interpretation that builds in these facts, see \cite{Fleisher}.} Rather than attempting to understand interpretability results directly, one could invoke the safety condition and use the interpretability results to search for nearby counterfactual situations where the model would have predicted incorrectly. Here, the idea is that if a model is making its decision based on an irrelevant feature, then the model risks making the wrong decision on a counterfactual input that lacks the target feature but possesses the irrelevant feature. If this is the case, then the model violates safety, since for a given data point, the situation where the irrelevant feature is present but the target feature is not is a similar possible world where the model predicts falsely. \cite{Degrave} consider such counterfactual inputs, using interpretability results for Covid detection to generate counterfactual images where healthy lungs are paired with background features from Covid patients and finding worse model performance on these counterfactual images. Such results suggest that the model can make unsafe decisions, getting the right answer in a way that is not robust to hypothetical changes to irrelevant background conditions.

Another technique in machine learning used to identify the reliance on the wrong features is causal shortcut detection \citep{Geirhos}. A machine learning algorithm relies on a causal shortcut when its output is based on a background feature that is correlated with, but causally disconnected from, the feature of interest. Common examples of causal shortcuts are background features in animal identification, such as the presence of water when detecting gulls and warblers \citep{Makar}. Causal shortcuts also pose a concern for algorithmic fairness, where variables like gender, race, or age may arise as undesirable shortcuts \citep{Makar1} as well as in text classification, where things like movie genre or use of technical language can serve as shortcuts for predictions like movie review sentiment \citep{Veitch1, Veitch, Zhao}.

Causal shortcut detection is similarly related to safety in that it can provide counterfactual data points relevant for determining if a model is safe. If one assumes that model inputs are generated by some causal process where the target feature is separate from other background features, then one can generate counterfactual data points by changing the target feature while leaving background features constant.\footnote{The connection between causal shortcut detection and safety is especially clear for the causal version of safety developed in \cite{Vandenburgh}.} For example, one can think of animal images as being generated by combining animals with an independently chosen background.\footnote{Determining how to model the data generating process is a difficult problem for causal shortcut detection. For example, animals and background are not independent of each other, so one may think that the data generating model should include the causal influence of the background on the distribution of animals. Some reasons for treating the two as independent may be that one can intervene on each variable separately or that the causal relationship is substantially confounded.} Thus, if a model correctly identifies a gull in a water background, one can test whether the identification is independent of the background as a causal shortcut by seeing whether the model would falsely predict that a warbler in the same background is a gull. If the model relies on a causal shortcut in this way, the identification is unsafe, since there is a similar input where the gull is changed out with a warbler and the model predicts falsely.

Both interpretability and causal shortcut detection can be thought of as attempts to ensure that models are making decisions based on the right features. Safety offers one way to map these approaches onto a counterfactual condition for reliance on the right features, where safe prediction requires that a model output cannot be triggered by background features or a causal shortcut in an input where the target feature is absent. This is a counterfactual requirement that depends on how well the model performs on hypothetical data points that one may never see in practice, and thus cannot be captured by just looking at model reliability. For example, a model relying on the presence of water or land to identify gulls and warblers could be very reliable in the real world, as this correlation often holds, but would fail when looked at through the lens of interpretability, causal shortcut detection, or safety.

Despite this connection, safety is weaker than some formulations of interpretability or causal shortcut independence. In particular, safety does not require that an interpretation will only highlight relevant features or that intermediate model representations will be independent of causal shortcuts. Consider a model to identify whether $x$ is 0 or 1 given binary $x$ and $y$, where the model calculates $z=x+0.3xy$ and outputs $x=1$ iff $z>0.5$. For input $(x,y)=(1,1)$, a local explanation of the model's output would reasonably indicate that both $x$ and $y$ were relevant factors, and the model's calculation of $z$ depends on $y$, which is causally independent of $x$ and therefore ought to be irrelevant. However, the model's predictions are safe, as even if $y$ had been 0, the model would have still predicted correctly -- in fact, the model always predicts correctly. This raises further questions about the role of intermediate steps in evaluating model behavior and about the nature of model explanations for binary prediction \citep{Kasirzadeh, Beisbart} and highlights how, if one is concerned with knowledge and safety, interpretation or shortcut detection may only offer defeasible evidence that something is going wrong.

\section{Distribution Shifts and Adversarial Robustness}

Safety requires that machine learning models produce the right answer for the right reason, offering an epistemic aim that is closely related to interpretability and causal shortcut detection. Another class of concerns raised for machine learning models is poor generalization, or lack of robustness to shifts in the data distribution. For example, suppose one has a Covid detector that is reliable at differentiating Covid-positive and healthy patients, but falsely identifies SARS as Covid. While this algorithm may be reliable and safe in contexts where the number of SARS cases is very low, if one deploys this model in a hospital where SARS is relatively prevalent, the model will perform poorly. Furthermore, the model's Covid diagnoses will all be unsafe in this context: even for a Covid patient that the algorithm diagnoses correctly, that patient could easily have been a SARS patient instead of a Covid patient, and in this similar scenario the algorithm would have been wrong. 

In this case, the distribution of patients in deployment is shifted from the training/testing distribution and from the normal deployment distribution to include more SARS patients. This distribution shift makes it such that possible worlds where the model diagnoses a SARS patient with Covid are no longer far away from the actual world. To see why this might be the case, we can think of the actual patient as being sampled from a local probability distribution. When SARS is relatively prevalent, the probability that a patient sampled from this distribution has SARS conditional on the model outputting a Covid diagnosis is relatively high, whereas this probability is low in standard contexts. It is plausible that the nearby worlds where the model outputs Covid include any possibilities that have a probability of obtaining above a certain threshold, so that having a SARS patient becomes a nearby possibility when the prevalence of SARS relative to Covid becomes sufficiently high.\footnote{\cite{Vandenburgh} defends such a conditional probability interpretation. For a more thorough discussion of similar cases, see \citet{Gendler}.}

These kinds of cases are not unique to machine learning and frequently arise in philosophical discussions of knowledge. For example, taking a hallucinogenic drug can shift the number of accurate perceptual experiences we have to the point where our perceptions are no longer reliable signals of reality. One of the most famous cases of a shifted data distribution for the theory of knowledge is Goldman's (\citeyear{Goldman1}) hypothetical fake barn case, where someone drives through an area with many papier-m\^{a}ch\'{e} barn fa\c{c}ades that look exactly like real barns. Even though the driver's perception is reliable in general, and even in the case where he is looking at one of the few real barns in the area and therefore has the right belief for the right reason, the high chance of misidentifying a barn fa\c{c}ade as a real barn undermines the possibility of knowledge. While these cases of distribution shift for human judgment may seem far-fetched, the rise of generative AI could provide real cases of such a shift by flooding our informational environment with fabricated content. For example, if it becomes sufficiently likely that a news story or photograph encountered in a certain environment (say, on social media) is fake, then we can no longer know that any of the stories or photographs are genuine without some further evidence for their reality.

While such epistemically unfriendly distribution shifts rarely occur for standard sources of knowledge like perception and testimony, they are a substantial concern for machine learning models. For example, models that are sensitive to background features picked up in training (such as the patient population, hospital procedures, or an animal's typical habitat) may fail when these background features shift, as in a new hospital system or when identifying animals in a zoo or someone's backyard \citep{Koh}. Additionally, work on performative prediction highlights how people may change their behavior based on knowledge about how an algorithm works, thereby shifting the data distribution and potentially undermining the algorithm's performance \citep{Perdomo}. In some cases, these failures may be widespread enough to undermine the overall reliability of the algorithm, but in other cases, these failures will only pose problems in certain deployment contexts. In these local cases, like with Goldman's fake barn case, the problem is not with the reliability of the belief-forming process as a whole, but rather with one's ability to obtain knowledge on this basis due to the nearby chance of error.

Another type of distribution shift posing concerns for machine learning algorithms is when interested parties change the data distribution to intentionally manipulate algorithmic outputs. Especially concerning is the possibility of adversarial attacks on model inputs, since research shows that there are many ways to manipulate an algorithm's input in small or imperceptible ways to trigger a desired output \citep{Carlini}. For example, \cite{Su} develop a method to change classification results by changing only one pixel in an input image. If adversarial attacks become common or likely enough in a given environment, this can shift the data distribution and undermine knowledge, since even if a data point is real, one could easily have obtained the same model output from a modified data point designed to falsely elicit this output. 

This concern that adversaries may shift an entire data distribution differs from the more commonly raised concern that machine learning models are not robust to adversarial attack. An algorithm is adversarially robust if one cannot trick the algorithm with an adversarial attack that changes the model's output with minimal or imperceptible changes to the input. Adversarial robustness is often seen as essential for ruling out problematic manipulations by adversaries and is sometimes connected to the desire to get the right answer for the right reason. For example, \cite{Ilyas} create adversarial examples where models make clearly absurd judgments, such as classifying an image that looks exactly like a horse as a dog.

I argue, however, that this stronger sense of adversarial robustness is not required for safety. In fact, human judgments are not adversarially robust, as there are always adversarial cases where human judgments fail. For example, when relying on testimony, we cannot always distinguish between the truth and a convincing lie, and a suitably motivated adversary could undermine our perception by dressing mules up like zebras \citep{Dretske} or by erecting barn fa\c{c}ades. The fact that these situations are possible in general does not undermine our ability to know based on testimony or perception in general; problems only arise in actual adversarial situations. In the terminology of possible worlds, these adversarial worlds are far away from the normal circumstances of the actual world, and are therefore not relevant for knowledge.\footnote{These adversarial cases are related to what \cite{Rathkopf} call ``strange errors" in machine learning. I take it that strange errors are not unique to machine learning, as falsely identifying barn fa\c{c}ades as barns seems strange when considering the world at the level of its material composition. Thus, the significance of these errors to machine learning lies not in their nature, but in the probability of encountering them.} Applying this to adversarial examples in machine learning, what matters is not whether an adversary is capable of tricking a model, but whether the model is deployed in an adversarial environment where such an attack is sufficiently likely to occur. While aiming for adversarial robustness offers one way of securing against deployment in an adversarial environment, this could also be accomplished by other methods like securing the system against adversary access.\footnote{Note, however, that the stronger sense of adversarial robustness may be required in domains where one cannot secure model inputs and where adversaries are likely to be sophisticated. For example, social media posts are very difficult to secure against adversarial manipulation in facial recognition or bot detection, and users and bots have an interest in developing increasingly sophisticated attacks to protect their identity. On the prevalence of bots on social media, see \cite{Varol}, and on adversarial attacks in facial recognition, see \cite{Shan}.} This also applies for robustness to distribution shifts in general, where what matters for our expectations of safety is not that a model be robust to all possible distribution shifts, but that the model be robust to the distribution shifts one expects to encounter in deployment \citep{Taori, Wu}.

\section{Conclusion}

I have argued that, in addition to being a reliable source of true belief, trustworthy algorithms should be a source of knowledge. To meet this standard, algorithms must produce the right answer for the right reason and work well in the local deployment context. Invoking the safety condition on knowledge, which requires that an algorithm's output could not have easily been false, suggests that many of the procedures developed to evaluate model robustness are closely connected to establishing knowledge. Poor interpretability results and dependence on causal shortcuts indicate that a model is making decisions for the wrong reasons. And a model that is not robust to likely shifts in the data distribution or is not secure against likely adversaries is liable to be deployed in an inappropriate context where it cannot ground knowledge. 

Knowledge matters insofar as it is often treated as the norm for believing, asserting, or acting  on the basis of something. However, it also matters for the ethics of algorithmic decision-making, since people can reasonably object to algorithms making decisions about them based on the wrong features or in a way that is not sensitive to their local situation. Thus, it is essential that those developing and deploying machine learning models justify not only their reliability, but also their robustness and potential for conferring knowledge. While the methods discussed above offer some tools for evaluating robustness, they do not exhaust the possibilities. For example, human-in-the-loop procedures can mitigate concerns about algorithms making mistakes that obviously rely on the wrong features or about algorithms failing catastrophically in a local context. New robustness techniques may also be needed for other machine learning tools such as generative language models, as current approaches may be inadequate for ensuring that generative outputs are grounded in the right causal features rather than untrustworthy sources or hallucinations \citep{Ji}.

\bibliography{preprint}{}
\bibliographystyle{agsm}
\end{document}